\pdfoutput=1

\documentclass[11pt]{article}

\usepackage[final]{acl}

\usepackage{times}
\usepackage{latexsym}

\usepackage[T1]{fontenc}

\usepackage[utf8]{inputenc}

\usepackage{microtype}

\usepackage{inconsolata}

\usepackage{paralist}

\usepackage{amsmath}
\usepackage{amssymb}
\usepackage[capitalise]{cleveref}

\usepackage[table]{xcolor}
\definecolor{lightblue}{RGB}{220,235,250}
\definecolor{lightgray}{RGB}{245,245,245}

\usepackage{xtab}
\usepackage{graphicx}
\usepackage{xspace}
\usepackage{booktabs}
\usepackage{multirow}
\usepackage{geometry}
\usepackage{array}
\usepackage[most]{tcolorbox}
\usepackage{caption}
\usepackage{adjustbox}
\usepackage{threeparttable}
\usepackage{threeparttablex}
\usepackage{arydshln}
\usepackage{svg}
\usepackage{subcaption}
\usepackage{mathtools}

\renewcommand{\paragraph}[1]{\noindent\textbf{#1}}

\tcbset{promptstyle/.style={
  colback=gray!5!white,
  colframe=gray!75!black,
  fonttitle=\bfseries,
  sharp corners,
  boxrule=0.8pt,
  enhanced,
  breakable,
}}

\newcommand{\eg}{e.g.,\xspace}

\title{Enhancing Long Document Long Form Summarisation \\
with Self-Planning}

\author{
\normalsize
Xiaotang Du$^1$\quad
Rohit Saxena$^1$\quad
Laura Perez-Beltrachini$^1$\\
\normalsize
\textbf{Pasquale Minervini}$^{1,2}$\quad
\textbf{Ivan Titov}$^1$
\\ 
\normalsize
$^1$University of Edinburgh, United Kingdom \qquad
$^2$Miniml.AI, United Kingdom\\
\normalsize
\texttt{\{xiaotang.du, p.minervini, ititov\}@ed.ac.uk}
}

\begin{document}
\maketitle
\begin{abstract}
We introduce a novel approach for long context summarisation, \textit{highlight-guided generation}, that leverages sentence-level information as a content plan to improve the traceability and faithfulness of generated summaries. 
Our framework applies self-planning methods to identify important content and then generates a summary conditioned on the plan. 
We explore both an end-to-end and two-stage variants of the approach, finding that the two-stage pipeline performs better on long and information-dense documents.
Experiments on long-form summarisation datasets demonstrate that our method consistently improves factual consistency while preserving relevance and overall quality.
On GovReport, our best approach has improved ROUGE-L by 4.1 points and achieves about 35\% gains in SummaC scores.
Qualitative analysis shows that highlight-guided summarisation helps preserve important details, leading to more accurate and insightful summaries across domains. \footnote{Our code is available at \url{https://github.com/acDante/llm-self-planning}}

\end{abstract}

\section{Introduction} \label{sec:introduction}

Despite the strong text generation capabilities of current large language models (LLMs), generated long-form summaries often diverge significantly from human references in both content and style~\citep{saxena2025endtoendlongdocumentsummarization}. 
When prompted for conciseness and relevance, LLMs frequently fail to operationalise these, i.e., they struggle to identify key information and remove unnecessary details. Moreover, their outputs are prone to hallucinations~\citep{askari-etal-2025-assessing,belem-etal-2025-single,DBLP:journals/tacl/ChrysostomouZWA24}.

Planning based approaches have been proposed to improve content selection (both in terms of saliency and coverage) as well as faithfulness in summarisation.
Most of these approaches rely on complex intermediate plans of different granularity such as entity chains \cite{DBLP:journals/tacl/NarayanZMSNM21}, keyphrases~\citep{DBLP:conf/emnlp/XuKDE24}, question-answer pairs~\citep{DBLP:journals/tacl/NarayanMAGLH00L23}, events~\citep{DBLP:journals/corr/abs-2504-09071}, discourse relations~\citep{DBLP:journals/corr/abs-2504-19339}, and topic templates \cite{perez-beltrachini-etal-2019-generating}.
In long-document summarisation, content selection is often implemented through an extract-then-generate pipeline, where sentences are selected using a trained classifier~\citep{DBLP:conf/emnlp/LiuL19,DBLP:conf/acl/OuL25} or similarity heuristics~\citep{lexrank}.

In this work, we argue that LLMs possess enough knowledge to identify key information in input documents to make their own plans. We propose a simple and effective approach without training based on self-planning, \textit{highlight-guided generation} (\textsc{HiGen}). We instruct LLMs to generate a summary together with its plan, i.e., a set of sentences highlighting salient content from the input document to support the generation of a summary.
We study two self-planning approaches. One where the sentence highlights are 
generated along with the summary (End-to-end) and a revision-based (Two-stage) one where the sentence highlights are fed back to the model together with the input document to generate a refined summary based on the highlights.  

An alternative self-planning approach can be implemented with attribution methods, which identify parts of the input that the LLM relies on when generating summaries.
We compare planning based on generative highlights versus planning based on extractive attribution methods.
Concretely, we compare with a perturbation-based attribution method that extracts those input document sentences that yield a decrease in summary quality when they are removed from the input. Generated highlights offer key advantages over attribution-based methods: they preserve contextual coherence (\eg maintaining speaker-utterance relationships in dialogues), are computationally more efficient than perturbation-based approaches, and can synthesise information rather than just extract sentences.

We evaluate our approach on two long-document and long-form summarisation datasets, including GovReport \cite{DBLP:conf/naacl/HuangCPJW21} and QMSum \cite{DBLP:conf/naacl/ZhongYYZMJACLQR21}, and measure summary quality in terms of relevance and faithfulness. 
Our experiments and analysis show that self-planning can effectively improve the overall quality of the generated summaries by 
enumerating summary worth points.

\section{Method} \label{sec:method}

\begin{figure*}[htbp]
\centering
\begin{subfigure}{0.4\textwidth}  %
    \centering
    \includegraphics[width=\textwidth]{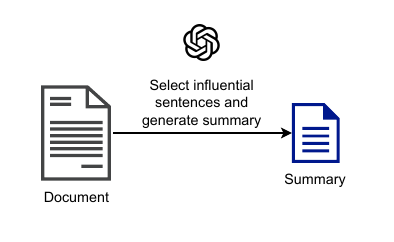}
    \caption{End-to-end}
    \label{fig:one-step}
\end{subfigure}%
\hspace{0.5cm}  %
\begin{subfigure}{0.45\textwidth}  %
    \centering
    \includegraphics[width=\textwidth]{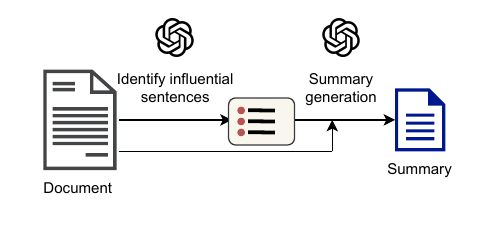}
    \caption{Two-stage}
    \label{fig:two-step}
\end{subfigure}
\caption{Illustration of highlight-guided generation framework for summarisation. The generated summary is grounded by the influential sentences extracted by the same model architecture.}
\label{fig:method}
\end{figure*}

We propose a novel self-planning summarisation framework for long-document summarisation that leverages sentence plans derived from the input document to guide the summary generation. 
Our approach is motivated by the observation that while LLMs possess sufficient knowledge to identify relevant content in input documents, they struggle with maintaining focus and avoiding hallucination in long-context scenarios.
By explicitly extracting important sentences as an intermediate content planning step, we aim to improve both the factual consistency and relevance of generated summaries.

Given an input document $D = \{s_1, s_2, \ldots, s_n\}$ consisting of $n$ sentences, our goal is to generate a summary $S$ that is both faithful to the source content and covers the most important information.
Traditional approaches directly map $D \rightarrow S$, while we introduce an intermediate content planning step by first identifying a subset of important sentences $H = \{h_1, h_2, \ldots, h_k\} \subseteq D$ where $k \ll n$, and then generating the summary conditioned on these highlights: $D \rightarrow H \rightarrow S$.

Our framework consists of two main components:
\begin{inparaenum}[(i)]
\item \emph{highlight generation}, which identifies the most important sentences from the input document, and
\item \emph{highlight-guided summarisation}, which generates the final summary based solely on the extracted highlights
\end{inparaenum}.
We explore two architectural variants that differ in how these components are integrated: an end-to-end approach that performs both steps in a single generation pass, and a two-stage pipeline that separates the highlighting and summarisation processes.

\subsection{End-to-end Approach}
In the end-to-end variant, we prompt the LLM to sequentially perform highlight extraction and summary generation within a single inference call.
The model is instructed to first identify and extract important sentences $H$ from the input document, then immediately generate a summary $S$ based only on the information contained in these highlights.

\subsection{Two-stage Pipeline}
To address the limitations of the end-to-end approach, we propose a two-stage pipeline that separates highlight extraction and summary generation into distinct inference calls.
In the first stage, the model extracts important sentences from the input document.
In the second stage, a fresh model context is used to generate the summary, with both the original document and the extracted highlights provided as input, but with explicit instructions to base the summary only on the highlighted content.
The two-stage process can be formalised as: $H = \text{LLM}(\text{prompt}_h, D)$ and $S = \text{LLM}(\text{prompt}_s, D, H)$, where $\text{LLM}(\cdot)$ denotes the language model inference function, $\text{prompt}_h$ and $\text{prompt}_s$ are the task-specific prompts for highlight extraction and summary generation respectively (see \cref{sec:appendix_prompt} for more details), $D$ is the input document, $H$ represents the extracted highlights, and $S$ is the final summary.

This separation offers several advantages: \begin{inparaenum}[(i)] \item it provides more reliable instruction following by focusing each generation step on a single task, \item it enables the use of different highlighting methods beyond generative extraction, and \item it allows for better control over the summary generation process by providing clear conditioning information\end{inparaenum}.

\subsection{Attribution Methods} \label{ssec:attribution}

Our two-stage framework supports multiple methods for extracting important sentences. A self-planning alternative to highlight
generation is context attribution.
Context attribution involves tracing and quantifying the influence of specific input segments on the generated output.
In this work, we investigate attribution methods that rely solely on the model's internal mechanisms.
We aim to investigate whether model attribution can effectively support content selection and guidance in long-context summarisation.

Perturbation-based methods quantify the importance of input sources by systematically perturbing the input and measuring the resulting changes in the model outputs, such as through occlusion~\citep{DBLP:conf/eccv/ZeilerF14,DBLP:conf/naacl/Ribeiro0G16,DBLP:conf/eacl/MohebbiZCA23,DBLP:journals/corr/abs-2402-00794,DBLP:conf/nips/Cohen-WangSGM24}.
In this work, we employ
ContextCite~\cite{DBLP:conf/nips/Cohen-WangSGM24}, a recently proposed context attribution method that identifies which parts of the input context most causally influence a model's generation by systematically ablating context elements and measuring the changes in output probabilities of the original response.

\section{Experiment Setting} \label{sec:experiment}

\paragraph{Datasets.}
We report the results on two long-form summarisation datasets from \textsc{Scrolls} benchmark \cite{DBLP:conf/emnlp/0002SIEYH0XGBL22}, including GovReport and QMSum. 

\paragraph{Evaluation metrics.}
We report several automatic metrics to assess various aspects of the generated summaries.
We use \textsc{Rouge-L} \cite{lin-2004-rouge} and \textsc{BERTScore-F1} \cite{DBLP:conf/iclr/ZhangKWWA20} to measure the \textit{relevance} of the summaries against human references.
We employ \textsc{SummaC} \cite{DBLP:journals/tacl/LabanSBH22} and \textsc{FactScore} \cite{DBLP:conf/emnlp/MinKLLYKIZH23} to assess the \textit{factual consistency} between the generated summary and input document. 
\textsc{SummaC} measures the overall consistency based on sentence-level entailment. Scores reported in the paper are computed using the \textsc{SummaC$_{Conv}$} model.
In this work, we adapt it to assess factual consistency by computing the percentage of atomic facts in the generated summary that are supported by the input document.
We use \texttt{gpt-4o-mini} model to compute \textsc{FactScore} values.
Additionally, we report the average length of the generated summary, measured in terms of the number of tokens.
We conduct paired \textit{t}-test to assess whether two metric values are significantly different at $p<0.05$.

\begin{table*}[t]
\centering
\resizebox{\textwidth}{!}{%
\begin{tabular}{l l c c c c c c c c c c}
\toprule
& & \multicolumn{5}{c}{\textsc{GovReport}} & \multicolumn{5}{c}{\textsc{QMSum}} \\
\cmidrule(lr){3-7} \cmidrule(lr){8-12}
\textsc{Model} & \textsc{Method} & \textsc{R-L} & \textsc{BS-F1} & \textsc{SumC} & \textsc{Fact} & \textsc{\#Tokens} & \textsc{R-L} & \textsc{BS-F1} & \textsc{SumC} & \textsc{Fact} & \textsc{\#Tokens} \\
\midrule
\multirow{6}{*}{\textbf{Llama3.1-8B}} & Direct & 45.49 & 62.08 & 53.48 & 78.80 & 532.08 & 21.99 & 57.22 & 36.61 & 71.62 & 139.58 \\
& LexRank & \hspace{0.5em}45.93$^{*}$ & 63.01 & 64.24 & 77.13 & 519.70 & 22.53 & 57.48 & \hspace{0.5em}37.11$^{*}$ & 70.93 & 136.68 \\
& SumCoT & 44.58  & 61.83 & 50.53 & \hspace{0.5em}86.27$^{*}$ & 545.61 & 22.49 & 56.14 & 38.05 & \textbf{81.79} & 131.71 \\
& \cellcolor{lightblue}HiGen-CC & \cellcolor{lightblue}44.12 & \cellcolor{lightblue}61.40 & \cellcolor{lightblue}\hspace{0.5em}59.77$^{*}$ & \cellcolor{lightblue}76.87 & \cellcolor{lightblue}479.18 & \cellcolor{lightblue}22.28 & \cellcolor{lightblue}57.30 & \cellcolor{lightblue}36.45 & \cellcolor{lightblue}69.75 & \cellcolor{lightblue}134.15 \\
& \cellcolor{lightblue}HiGen (end-to-end) & \cellcolor{lightblue}39.62 & \cellcolor{lightblue}\hspace{0.5em}62.14$^{*}$ & \cellcolor{lightblue}59.23 & \cellcolor{lightblue}\textbf{88.45} & \cellcolor{lightblue}382.49 & \cellcolor{lightblue}\textbf{23.78} & \cellcolor{lightblue}\textbf{58.05} & \cellcolor{lightblue}\textbf{38.39} & \cellcolor{lightblue}\hspace{0.5em}77.20$^{*}$ & \cellcolor{lightblue}118.53 \\
& \cellcolor{lightblue}HiGen (two-step) & \cellcolor{lightblue}\textbf{47.18} & \cellcolor{lightblue}\textbf{63.08} & \cellcolor{lightblue}\textbf{65.68} & \cellcolor{lightblue}83.38 & \cellcolor{lightblue}566.41 & \cellcolor{lightblue}\hspace{0.5em}22.76$^{*}$& \cellcolor{lightblue}\hspace{0.5em}57.78$^{*}$ & \cellcolor{lightblue}36.51 & \cellcolor{lightblue}76.88 & \cellcolor{lightblue}142.38 \\
\midrule
\multirow{6}{*}{\textbf{Qwen3-8B}} & Direct & \hspace{0.5em}43.08$^{*}$ & \hspace{0.5em}63.35$^{*}$ & 47.97 & 89.99 & 491.42 & 22.56 & 58.43 & 37.88 & 81.57 & 121.53 \\
& LexRank & 46.79 & 64.61 & \hspace{0.5em}59.02$^{*}$ & 88.65 & 649.58 & \hspace{0.5em}22.73$^{*}$ & \hspace{0.5em}58.44$^{*}$ & 37.97 & \hspace{0.5em}81.73$^{*}$ & 122.82 \\
& SumCoT & 34.19 & 60.66 & 43.92 & \hspace{0.5em}90.14$^{*}$ & 307.04 & 22.02 & 57.67 & \textbf{39.78} & 84.79 & 108.44 \\
& \cellcolor{lightblue}HiGen-CC & \cellcolor{lightblue}46.60 & \cellcolor{lightblue}64.46 & \cellcolor{lightblue}56.34 & \cellcolor{lightblue}88.34 & \cellcolor{lightblue}639.04 & \cellcolor{lightblue}\textbf{23.38} & \cellcolor{lightblue}\textbf{58.74} & \cellcolor{lightblue}37.49 & \cellcolor{lightblue}79.42 & \cellcolor{lightblue}127.24 \\
& \cellcolor{lightblue}HiGen (end-to-end) & \cellcolor{lightblue}38.75 & \cellcolor{lightblue}58.67 & \cellcolor{lightblue}46.54 & \cellcolor{lightblue}85.70 & \cellcolor{lightblue}407.37 & \cellcolor{lightblue}22.35 & \cellcolor{lightblue}58.39 & \cellcolor{lightblue}39.39 & \cellcolor{lightblue}85.01 & \cellcolor{lightblue}105.14 \\
& \cellcolor{lightblue}HiGen (two-step) & \cellcolor{lightblue}\textbf{47.20} & \cellcolor{lightblue}\textbf{64.71} & \cellcolor{lightblue}\textbf{65.73} & \cellcolor{lightblue}\textbf{91.07} & \cellcolor{lightblue}709.32 & \cellcolor{lightblue}22.94 & \cellcolor{lightblue}58.71 & \cellcolor{lightblue}\hspace{0.5em}38.23$^{*}$ & \cellcolor{lightblue}\textbf{86.28} & \cellcolor{lightblue}131.39 \\
\midrule
\multirow{5}{*}{\textbf{Qwen3-32B}} & Direct & 43.19 & 63.59 & 48.14 & \textbf{90.33} & 453.16 & 21.76 & 58.25 & 37.01 & 80.02 & 132.22 \\
& LexRank & \hspace{0.5em}45.07$^{*}$ & \textbf{64.73} & \hspace{0.5em}56.19$^{*}$ & \hspace{0.5em}87.67$^{*}$ & 533.38 & 21.91 & \textbf{58.38} & 37.12 & 79.62 & 130.24 \\
& SumCoT & 36.40 & 61.90 & 43.48 & 89.96 & 326.73 & 22.27 & 57.84 & \hspace{0.5em}37.44$^{*}$ & \hspace{0.5em}80.39$^{*}$ & 125.50 \\
& \cellcolor{lightblue}HiGen (end-to-end) & \cellcolor{lightblue}45.80 & \cellcolor{lightblue}63.78 & \cellcolor{lightblue}50.23 & \cellcolor{lightblue}86.41 & \cellcolor{lightblue}570.81 & \cellcolor{lightblue}\textbf{22.37} & \cellcolor{lightblue}58.24 & \cellcolor{lightblue}\textbf{38.25} & \cellcolor{lightblue}76.21 & \cellcolor{lightblue}119.96 \\
& \cellcolor{lightblue}HiGen (two-step) & \cellcolor{lightblue}\textbf{46.39} & \cellcolor{lightblue}\hspace{0.5em}64.14$^{*}$ & \cellcolor{lightblue}\textbf{60.82} & \cellcolor{lightblue}89.98 & \cellcolor{lightblue}619.77 & \cellcolor{lightblue}\hspace{0.5em}21.62$^{*}$ & \cellcolor{lightblue}\hspace{0.5em}58.01$^{*}$ & \cellcolor{lightblue}36.88 & \cellcolor{lightblue}\textbf{83.36} & \cellcolor{lightblue}144.40 \\
\bottomrule
\end{tabular}
}
\caption{Model performance on GovReport and QMSum validation sets measured in terms of \textsc{ROUGE-L}, \textsc{BERTScore-F1}, \textsc{SummaC}, \textsc{FactScore}, and summary length in tokens. Performance of \textsc{HiGen} variants are \colorbox{lightblue}{highlighted}. \textsc{HiGen-CC} denotes \textsc{HiGen} pipeline with highlights extracted by ContextCite attribution. \textbf{Bold} indicates best results for each metric. We also mark with $^{*}$ the next best values that are significantly different based on paired \textit{t}-test ($p<0.05$).}
\label{tab:combined_performance}
\end{table*}

\paragraph{Models and Baselines.}
We evaluate the performance of \texttt{Llama3.1-8B}, \texttt{Qwen3-8B} and \texttt{Qwen3-32B} on the long-form summarisation datasets, with and without self-planning.
The experiments are conducted in a zero-shot setting.
As baselines for comparison, we consider direct prompting, Summary Chain-of-Thought~\cite[SumCoT,][]{wang-etal-2023-element} and extractive summariser LexRank \cite{lexrank}.
In the direct prompting setting, the model directly generates the summary given the input document, without content selection steps. 
SumCoT is a two-stage pipeline that leverages a QA-based plan to guide the summarisation process.
LLMs are instructed to extract important information about entities and events by answering a list of guiding questions and then produce a summary with more fine-grained detail by integrating extracted information. 
We also compare the performance of HiGen with a two-step pipeline that combines LLM with sentence extraction by an external unsupervised module. Concretely, we extract key sentences using graph-based saliency scoring with LexRank \cite{lexrank} and then prompt the LLM to generate the summary based on the selected content.

\paragraph{Hyperparameters.}
When generating the summaries, we apply greedy decoding with a temperature of 0 to produce deterministic outputs.
On both GovReport and QMSum dataset, we extract 30 influential sentences as the highlights for LexRank baseline and all the HiGen variants.

\section{Results} \label{sec:discussion}
Results in \cref{tab:combined_performance} highlight consistent trends that demonstrate the effectiveness of our attribution-guided summarisation approach in long-context summarisation.
Compared with the baselines, the proposed two-step pipeline significantly enhances both relevance and factual consistency for all models considered in the experiments on GovReport. 
For example, with \texttt{Qwen3-8B} model, the attribution-guided approach helps improve \textsc{ROUGE-L} from 43.08 to 47.20, indicating that the generated summaries more closely align with the content covered in the human references.
Meanwhile, \textsc{SummaC} increases from 47.97 to 65.73 and \textsc{FactScore} improves from 0.8999 to 0.9107, which suggests the generated summaries are more faithfully supported by the input document.
Our two-stage pipeline consistently outperforms the end-to-end approach on the GovReport dataset, showing significantly better \textsc{ROUGE-L} and \textsc{SummaC} scores, while the two model variants show comparable performance on the QMSum dataset.
This result implies that a separate content selection step benefits more in complex documents with dense information.

\paragraph{Generative highlights achieve a better balance between relevance and faithfulness.}
We compare the performance of generative highlights against LexRank and attribution-based highlights in our summarisation framework.
Appendix \cref{tab:lexrank_attr_sample} elaborates the difference between generative highlights and highlights extracted by LexRank. 
LexRank tends to capture the high-level structure of the document, often selecting shorter sentences or section headers as highlights, while generative approach produces more informative highlights.

For the experiments with ContextCite attribution, we only take into account important sentences with non-zero attribution scores when generating the summary. %
\cref{tab:combined_performance} shows that the relevance of the summaries guided by ContextCite attribution is comparable to the summaries guided by generative highlights.
Generative highlights consistently outperform ContextCite attribution in terms of factual consistency, showing better \textsc{SummaC} and \textsc{FactScore} scores across different models and datasets. 
Example summaries in \cref{tab:qmsum_sample} demonstrate that ContextCite attribution encourages the model to produce a comprehensive summary that is rich in detail, including both the final decisions and specific action items.
The summary can be lengthy and verbose compared with the summaries guided by generative highlights.

\cref{tab:attr_sample} shows that in the meeting summarisation task, many highlights extracted by ContextCite attribution are not informative, while the generative highlights are able to extract the key facts by synthesising information in the local context.
Generative highlights can preserve the speaker-utterance correspondence in the dialogue by converting the utterance into a concise statement.

\paragraph{Qualitative analysis.}
We conduct a qualitative analysis on 20 examples drawn from each of the GovReport and QMSum datasets.
As shown in the examples in the \cref{sec:qualitative_analysis}, the baseline summaries often capture a broad and high-level overview of the meeting, with a focus on the key decisions made during the meeting, rather than addressing the specific query.
While SumCoT summaries provide a consistent and clear structure that involves meeting attendees, discussion topics and main decisions, speaker attribution, concrete technical detail and the rationale behind decisions are often omitted in the summaries.
Summaries guided by the highlights include not only the final decisions, but also the core rationale and trade-offs behind them.
Our analysis reveals that highlight-guided summarisation can help preserve important details, such as entities, terminology, and quantitative data, which are often omitted in summaries generated by direct prompting.
Our technique also proves beneficial in query-based summarisation, where the model leverages the extracted highlights to identify relevant information and generate more targeted and query-aligned summaries.

\section{Related Work}

To deal with the length of the input document, existing summarisation approaches implement an
extractive-abstractive pipeline. The extractive step performs explicit content selection that serves as a plan of what should be included in the summary.
Some approaches rely on unsupervised sentence extraction methods \cite{lexrank, DBLP:journals/corr/BarriosLAW16, DBLP:conf/acl/ZhengL19,padmakumar-he-2021-unsupervised}, while others require the
learning of a task-specific content planning module.
This module often operates at different levels of granularity, including entity chains \cite{DBLP:journals/tacl/NarayanZMSNM21}, keyphrases \cite{DBLP:conf/emnlp/GehrmannDR18, DBLP:conf/emnlp/XuKDE24}, sentences \cite{DBLP:conf/emnlp/SharmaHHW19,liu-lapata-2019-hierarchical,lebanoff-etal-2019-scoring}, topic templates \cite{perez-beltrachini-etal-2019-generating}, question-answer pairs \cite{DBLP:journals/tacl/NarayanMAGLH00L23}, and discourse relations \cite{DBLP:journals/corr/abs-2504-19339}.
Following this paradigm, our method also integrates sentence-level content planning into the summarisation pipeline.
Unlike previous planning-based approaches, it does not require the annotation of content plans nor training a separate content planning module.

With recent advances in LLMs' long context understanding ability, an increasing number of studies have explored prompting-based approaches \cite{wang-etal-2023-element, DBLP:conf/emnlp/XuKDE24} to plan the content of the summary.
SumCoT \cite{wang-etal-2023-element} instructs LLMs to generate summaries step-by-step via answering a set of guiding questions. In multi-document and long-form summarisation, compressing long context into a selection of key points for hierarchical processing is a widely used approach \cite{bhaskar-etal-2023-prompted, DBLP:conf/naacl/HuangLFCJXW24, padmakumar-etal-2025-principled, kim2025nexussumhierarchicalllmagents} .
Similar to prior work, our method extracts influential sentences from the document by prompting.
However, it is more generalisable across different domains, without introducing task-specific prompt design.

\section{Conclusions}
\label{sec:conclusion}
We introduced an highlight-guided summarisation framework for long-document summarisation that leverages important sentence-level information to improve both factual consistency and relevance of generated summaries.
Our approach addresses key challenges in long-context summarisation by explicitly identifying important content before generation, mimicking human summarisation processes.
Our experiments on GovReport and QMSum demonstrate consistent improvements across multiple models. The two-stage pipeline achieves substantial gains in ROUGE-L scores (up to 4.1 points on GovReport) and factual consistency metrics, with SummaC scores improving from 48.0 to 65.7.
Our qualitative analysis reveals that the proposed framework can help preserve important details such as entities, terminology, and quantitative data that are often omitted in direct prompting approaches.

\section*{Limitations}
This study examines the potential of LLM attribution to enhance content selection and proposes an effective approach for long-document summarisation based on self-planning.
Despite the promising experimental outcomes, we acknowledge several limitations of the framework that we aim to address in future work.
Firstly, the computational overhead of the two-stage pipeline increases inference time and resource requirements compared to direct prompting, particularly when using perturbation-based attribution methods.
The generative highlighting approach, while effective for summarising dialogues and scattered information can 
exhibit positional bias in content coverage inherent to the underlying LLM.
Variations in prompt instructions may lead to different extracted sentences.
We optimise the prompt templates manually, without performing a systematic analysis of prompt sensitivity.
Moreoever, our experiments are limited to single-document summarisation, where the document length can fit within the context window of the considered LLMs.
The main reason for this choice is that perturbation-based attribution methods are computationally expensive for very long input context.
With generative highlights, the same principle can be extended to long narratives or multiple documents by integrating it with hierarchical or iterative frameworks to refine content selection across multiple stages.

\section*{Acknowledgements}
We thank the anonymous reviewers for the insightful feedback and comments.
We also thank Sherrie Shen and Ye Wang for their valuable support with human evaluation and review.
We are grateful to Miao Li and Zheng Zhao for helpful discussions.
Xiaotang Du was partly supported by the UKRI Centre for Doctoral Training in Natural Language Processing, funded by UK Research and Innovation (grant EP/S022481/1) and the University of Edinburgh, School of Informatics.
Pasquale Minervini was partially funded by ELIAI (The Edinburgh Laboratory for Integrated Artificial Intelligence), EPSRC (grant no.\ EP/W002876/1), an industry grant from Cisco, and a donation from Accenture LLP.
Laura Perez-Beltrachini was supported by the UK Engineering and Physical
Sciences Research Council (grant EP/W002876/1).
This work was supported by the Edinburgh International Data Facility (EIDF) and the Data-Driven Innovation Programme at the University of Edinburgh.

\bibliography{custom}

\clearpage

\appendix

\section{Experimental Setup Details}
\subsection{Dataset Details}
The licenses for the datasets used in our experiments are as follows.
QMSum is available under MIT License, and the original GovReport dataset is available under CC-BY-4.0 License.
For both datasets, we use the version from SCROLLS benchmark, which is under MIT License.
We run experiments on 300 samples from the GovReport validation set.
Experiments on QMSum are run on the whole validation set (272 samples).

\subsection{Implementation Details}
We implement \textsc{ROUGE-L} and \textsc{BERTScore} using \texttt{evaluate} library. \textsc{BERTScore} is computed by \texttt{DeBERTa-xlarge-mnli model} \cite{he2021deberta}, 
We adopt the implementation of \textsc{FactScore} from PRISMA code repository \cite{DBLP:conf/acl/MahonL24} and use \texttt{GPT-4o-mini} for both atomic fact extraction and claim verification.
We adapt the implementation of ContextCite from \cite{DBLP:conf/nips/Cohen-WangSGM24} to extract ContextCite attributions.
Input sentences are ranked by their attribution scores, and the top-$k$ sentences are selected as highlights.
We extract 30 attributed sentences for each instance.
We only take into account the important sentences with non-zero attribution scores when producing the summaries.

\section{Prompt Templates} \label{sec:appendix_prompt}
We present the prompt templates used for highlight extraction and summary generation in this section.
\cref{prompt_highlight_extraction_gr} and \cref{prompt_gen_summary_gr} show the prompt templates used for the experiments on GovReport dataset.
We adapted the standard prompt used in LongBench \cite{DBLP:conf/acl/BaiLZL0HDLZHDTL24} and ZeroSCROLLS Benchmark \cite{DBLP:conf/emnlp/0002IEBL23} and added instructions to enforce structured output.
For each dataset, the model is instructed to generate summaries that match the average length of the reference summaries ton ensure fair comparison.

\cref{prompt_highlight_extraction_qmsum} and \cref{prompt_gen_summary_qmsum} show the prompt templates used for the experiments on QMSum dataset. 
The prompt format is adapted from \cite{wan-etal-2025-positional}.

\section{Computation Details}
Experiments with ContextCite attribution as highlights were run on four NVIDIA A100 GPUs with 80GB of GPU memory.
Other experiments were run on two NVIDIA A100 GPUs.
The GPU hours vary depending on the model size and average context length in the dataset.
Extracting ContextCite attribution using \texttt{Qwen3-8B} model on GovReport validation set takes about 10 hours.
Generating the highlights and summaries on QMSum or GovReport takes about 30 minutes to 1 hour with \texttt{vllm}. \cite{kwon2023efficient}

\section{Qualitative Examples}
\label{sec:qualitative_analysis}
This section provides qualitative examples of summaries generated by different methods and different types of highlights. 

\cref{tab:attr_sample} demonstrates the difference between the salient sentences extracted by ContextCtie attribution and highlight sentences generated by LLM on a random instance from QMSum validation set. 
Both ContextCite attribution and generated highlights are computed using \texttt{Qwen3-8B} model.

\begin{figure}[]
\centering
\begin{tcolorbox}[colback=gray!10!white,colframe=black!50!black,title=Hightlight Extraction + Summary Generation,fonttitle=\bfseries, halign title=flush center, width=0.5\textwidth]
{You are given a report by a government agency. Extract a list of \{\texttt{Number of sentences}\} key sentences from the input document and then write a one-page summary of the report only focusing on the extracted sentences. You must give your answer in a structured format: "Key Sentences:\\
1. \{\texttt{Sentence Text}\} \\
2. \{\texttt{Sentence Text}\} \\
... \\
Summary: [your summary]", where [your summary] is your generated summary.\\
\\
Report: \\
\{\texttt{Document Text}\}}
\end{tcolorbox}
\caption{Prompt used for end-to-end highlight extraction and summary generation on GovReport}
\label{prompt_highlight_extraction_gr}
\end{figure}

\begin{figure}[]
\centering
\begin{tcolorbox}[colback=gray!10!white,colframe=black!50!black,title=Summary Generation,fonttitle=\bfseries, halign title=flush center, width=0.5\textwidth]
{You are given a report by a government agency. Write a one-page summary of the report focusing on the main points. You must give your answer in a structured format: "Summary: [your summary]", where [your summary] is your generated summary.\\
\\
Report:\\
\{\texttt{Document Text}\}\\
You should only focus on the following key points:\\
1. \{\texttt{Sentence Text}\} \\
2. \{\texttt{Sentence Text}\} \\
...
}
\end{tcolorbox}
\caption{Prompt used for generating the summary with the two-step pipeline on GovReport.}
\label{prompt_gen_summary_gr}
\end{figure}

\begin{figure}[]
\centering
\begin{tcolorbox}[colback=gray!10!white,colframe=black!50!black,title=Hightlight Extraction + Summary Generation,fonttitle=\bfseries, halign title=flush center, width=0.5\textwidth]
{Read the following meeting transcript. Extract a list of \{{\tt Number of sentences}\} key sentences from the input document and then produce a summary in 4 sentences only focusing on the extracted sentences. You must give your answer in a structured format: "Key Sentences:\\
1. \{{\tt Sentence Text}\} \\
2. \{{\tt Sentence Text}\} \\
... \\
Summary: [your summary], where [your summary] is your generated summary.
========== \\ 
\{{\tt Meeting Transcript}\} \\
==========
}
\end{tcolorbox}
\caption{Prompt used for end-to-end highlight extraction and summary generation on QMSum}
\label{prompt_highlight_extraction_qmsum}
\end{figure}

\begin{figure}[]
\centering
\begin{tcolorbox}[colback=gray!10!white,colframe=black!50!black,title=Summary Generation,fonttitle=\bfseries, halign title=flush center, width=0.5\textwidth]
{Read the following meeting transcript. Produce a summary in 4 sentences focusing on key decisions, action items, and important discussion points. You must give your answer in a structured format:
"Summary: [your summary]", where [your summary] is your generated summary.\\
========== \\ 
\{\texttt{Meeting Transcript}\} \\
========== \\
You should only focus on the following key points:\\
1. \{\texttt{Sentence Text}\} \\
2. \{\texttt{Sentence Text}\} \\
...
}
\end{tcolorbox}
\caption{Prompt used for generating the summary with the two-step pipeline on QMSum.}
\label{prompt_gen_summary_qmsum}
\end{figure}

\begin{table*}[t]

\centering
\footnotesize
  \renewcommand\arraystretch{1}
  \setlength{\tabcolsep}{2.5mm}{

\begin{xtabular}{p{1.98\columnwidth}}

\toprule

\textbf{Gold Summary}\\
\midrule

The industrial designer recommended batteries instead of solar cells because it would be problematic if solar cells didn't have enough energy. Titanium was proposed rather than rubber as the latter would make it difficult to do the moulding of buttons. Besides, the industrial designer wanted a simple push button because it only needed a simple chip and was less expensive than LCD. \\

\midrule
\textbf{Baseline Summary}\\
\midrule

The industrial designer recommended using simple batteries instead of solar cells and titanium for the remote control's material due to manufacturing and cost concerns. They also suggested push buttons over LCD displays to reduce cost and complexity, while acknowledging the possibility of an LCD for information output if budget allows. The team discussed incorporating trendy, spongy, and fruit-like designs, considering both titanium and rubber materials, and explored the idea of making the remote control customizable with additional features like color options or speech recognition. The final decision was to focus on creating a fancy and innovative product that aligns with current design trends, even if it means sacrificing some functionality for aesthetic appeal. \\

\midrule
\textbf{SumCoT Summary}\\
\midrule

The document describes a conceptual design meeting involving the Project Manager, Marketing, User Interface, and Industrial Designer, focusing on the design of a remote control. Key discussions revolve around materials, interface options, and incorporating trend elements like a spongy, fruit-shaped aesthetic. The team decides to use simple batteries and push buttons instead of solar cells or LCD displays due to cost and manufacturing constraints. They plan to explore two design options—titanium with a fruit scent or a spongy rubber body—while keeping the product aligned with current trends and within budget. \\

\midrule
\textbf{Highlight-driven planning (End-to-end)} \\
\midrule

The industrial designer recommended using simple batteries and push buttons for the remote control due to cost and manufacturing constraints, while also considering the use of titanium for design. They discussed the feasibility of using a speech recognition chip to enhance functionality without increasing costs too much. The team also explored the idea of incorporating trendy, natural elements like fruit and vegetable shapes and colors, and considered optional features such as color-changing displays or fingerprint recognition. Ultimately, they decided to focus on creating a fancy, user-friendly remote control that aligns with current design trends.\\

\midrule
\textbf{Highlight-driven planning (Two-step, ContextCite)}\\
\midrule
The team discussed the design of industrial components for the remote control, deciding to use simple batteries instead of solar cells due to energy concerns, and opted for titanium over rubber or plastic for its aesthetic appeal, despite manufacturing challenges. They agreed to use push buttons rather than LCD displays to reduce costs and simplify the technology, while also considering the inclusion of speech recognition for added functionality. Marketing emphasized the importance of making the remote control \"fancy\" and highlighted the trend of using fruit and vegetable-inspired colors and materials to differentiate the product. The project manager assigned individual tasks for Rama, Mark, and Sammy, with the group planning to create a prototype using modeling clay and explore both titanium and spongy design options.\\

\midrule
\textbf{Highlight-driven planning (Two-step, Generative)}\\
\midrule

The industrial designer recommended using simple batteries for energy and titanium for the material due to manufacturing feasibility and cost-effectiveness, while avoiding double-curved shapes. They also suggested using push buttons instead of an LCD display to reduce costs and simplify technology, as the LCD requires more advanced and expensive components. The team discussed the possibility of incorporating speech recognition technology, which would allow for more features but needed to be balanced with budget constraints. The designer proposed exploring a spongy, fruit-shaped design with titanium panels and a rubber body, aiming to align with current trends and enhance the product's appeal. \\

\bottomrule 

\end{xtabular}}

\caption{Comparison between different summaries generated by \texttt{Qwen3-8B} model for a random sample from QMSum validation set. Summaries produced using highlight-driven planning are guided by 30 sentences.  }
\label{tab:qmsum_sample}%
    
\end{table*}

\begin{table*}[t]

\centering
\footnotesize
  \renewcommand\arraystretch{1}
  \setlength{\tabcolsep}{2.5mm}{

\begin{xtabular}{p{1.98\columnwidth}}

\toprule

\textbf{Gold Summary}\\
\midrule

Marketing suggested putting a small LCD screen on top of the remote but Industrial Designer against it because a remote should be started with the power button and User Interface thought it would be unnatural to put the LCD on the top. Then, Marketing mentioned that would be where users' handball might be. So Industrial Designer thought they could put the whole interface a bit down so there would be room for the interface. \\

\midrule
\textbf{Baseline Summary}\\
\midrule

The group discussed the pros and cons of including an LCD screen on the remote, emphasizing its importance for the younger target audience despite cost concerns. They agreed to prioritize the most frequently used functions like power, channel, and volume, placing them in a central and accessible location to reduce RSI issues. The design should be familiar yet modern, incorporating the company's corporate image and ensuring usability. Action items include evaluating the cost of LCD screens and finalizing the placement of the screen and buttons before the next design round. \\

\midrule
\textbf{SumCoT Summary}\\
\midrule

The meeting involves the Marketing team, Project Manager, Industrial Designer, and User Interface Designer discussing the design and functionality of a remote control. They are considering the inclusion of an LCD screen and debating its placement, as well as the layout and usability of buttons. The group agrees to include an LCD screen but needs to finalize its position and cost. They also decide to prioritize the most commonly used functions and focus on a younger audience, aiming for a user-friendly and brand-recognizable design. \\

\midrule
\textbf{Highlight-driven planning (End-to-end)} \\
\midrule

The group discussed the design and functionality of a remote control, emphasizing the need for a familiar yet modern look, with a focus on user-friendly features like an LCD screen. They considered the placement of buttons, aiming to reduce RSI by spacing out frequently used ones. The team also debated the importance of a recognizable corporate image and the inclusion of features like a beeping button to locate the remote. Despite the cost concerns, they agreed that an LCD screen is essential to meet the preferences of their target younger audience.\\

\midrule
\textbf{Highlight-driven planning (Two-step, ContextCite)}\\
\midrule
The group discussed the pros and cons of including an LCD screen on the remote, noting that while it is important for the younger audience and offers useful features like program recognition, it may be costly and not essential for all users. They agreed that the most important functions—power, channel, and volume—should be placed near the thumb to reduce the risk of RSI, but also considered spacing them apart to avoid repetitive strain. The Industrial Designer mentioned the need to consider the cost and design implications of the LCD screen, while the Marketing team emphasized the importance of a user-friendly design and the need to prioritize key features due to time constraints. The group ultimately decided to include an LCD screen but acknowledged the need to explore its feasibility and cost before finalizing the design.\\

\midrule
\textbf{Highlight-driven planning (Two-step, Generative)}\\
\midrule

The group discussed the pros and cons of including an LCD screen on the remote, noting its importance for the younger audience but acknowledging the cost concerns. They agreed that the remote should be designed to fit the hand, be original yet familiar, and incorporate a recognizable corporate image, possibly through a slogan or color. The decision was made to include an LCD screen, though its placement and functionality will be further discussed, with the focus on making it user-friendly and intuitive. The team also emphasized the need for a short learning curve, with the most frequently used buttons (power, channel, volume) placed for easy access and to reduce RSI issues. \\

\bottomrule 

\end{xtabular}}

\caption{Comparison between different summaries generated by \texttt{Qwen3-8B} model for a random sample from QMSum validation set. Summaries produced using highlight-driven planning are guided by 30 sentences.  }
\label{tab:qmsum_sample2}%
    
\end{table*}

\begin{table*}[t]

\centering
\footnotesize
  \renewcommand\arraystretch{1}
  \setlength{\tabcolsep}{2.5mm}{

\begin{xtabular}{p{1.98\columnwidth}}

\toprule

\textbf{Gold Summary}\\
\midrule

When the group found that it was hard to balance between the cost and functions, the Project Manager proposed to give up the voice recognition and keep the touch screen. But according to Marketing’s research, even if the interviewees were not informed of the touch screen, they showed a great preference for voice recognition. So just for catering to popular tastes, the Marketing insisted that they should keep the voice recognition in any case. \\

\midrule
\textbf{Baseline Summary}\\
\midrule

The team discussed the feasibility of incorporating a touch screen into the remote control design but ultimately decided to prioritize voice recognition due to its lower cost and existing in-house technology. They agreed to use rubber buttons and a non-traditional, organic design for the casing to align with current market trends and appeal to the youth market. A backup battery was suggested to complement solar power, ensuring reliability in low-light conditions. The next steps involve developing a prototype with a simple, non-remote-looking design, focusing on voice recognition and a sleek, versatile interface, with Nathan and Ron collaborating on the design and layout. \\

\midrule
\textbf{SumCoT Summary}\\
\midrule

The document describes a meeting between the Marketing team, Project Manager, Industrial Designer, and User Interface team to discuss the design of a new remote control. The main topic of debate is whether to use a touch screen or voice recognition technology, with the latter being chosen due to budget constraints. They also consider materials for the case, such as rubber or natural options, and aim for a simple, non-traditional design. The result is a decision to prioritize voice recognition and focus on a sleek, user-friendly remote control that aligns with current market trends. \\

\midrule
\textbf{Highlight-driven planning (End-to-end)} \\
\midrule

The meeting discusses the conceptual design of a remote control, focusing on the touch screen and voice recognition features. Marketing is concerned about the cost of the touch screen, while the Industrial Designer suggests it could be a selling point. They debate between touch screen and voice recognition, ultimately deciding to prioritize voice recognition due to its lower cost and existing in-house technology. The team also considers alternative designs and materials for the remote, aiming to create a unique and user-friendly product.\\

\midrule
\textbf{Highlight-driven planning (Two-step, ContextCite)}\\
\midrule
The team discussed the limitations of the manufacturer in component design, particularly regarding the touch screen and energy sources, and considered alternatives like solar or kinetic power. They decided to provisionally go with the touch screen as a main selling point but acknowledged the high costs involved. After evaluating the costs and market appeal, they agreed to prioritize voice recognition over the touch screen due to its fiscal viability and existing in-house technology. The Project Manager assigned Nathan to focus on the basic look and feel of the remote, while Ron was tasked with designing the layout and prototype, aiming to target the youth market with a sleek, non-traditional remote design.\\

\midrule
\textbf{Highlight-driven planning (Two-step, Generative)}\\
\midrule

The Marketing team disagreed with the Project Manager regarding the touch screen due to budget constraints, as it would significantly increase costs. The meeting focused on conceptual design, aiming to finalize decisions for the prototype and clarify roles. The Industrial Designer discussed component options, including energy sources and case materials, while highlighting the cost implications of touch screen technology. The team ultimately decided to prioritize voice recognition over touch screens to balance innovation with budget feasibility, and agreed to proceed with a simple, non-traditional remote design incorporating rubber and multiple case colors. \\

\bottomrule 

\end{xtabular}}

\caption{Comparison between different summaries generated by \texttt{Qwen3-8B} model for a random sample from QMSum validation set. Summaries produced using highlight-driven planning are guided by 30 sentences.  }
\label{tab:qmsum_sample3}%
    
\end{table*}

\begin{table*}[t]

\centering
\footnotesize
  \renewcommand\arraystretch{1}
  \setlength{\tabcolsep}{2.5mm}{

\begin{xtabular}{p{1.98\columnwidth}}

\toprule

\textbf{Gold Summary}\\
\midrule

Mental spaces can be tackled with mechanisms that can also deal with context issues (time, space etc.): creating a base space and rules of interaction with other interconnected spaces. However, the complexity of these mechanisms has to be bound as well: it is necessary to define the range of constructions to be studied. \\

\midrule
\textbf{Baseline Summary}\\
\midrule

The group discussed the updated semantic specification and construction formalism, noting that while the notation is still evolving, it aims to capture both form and meaning types. They agreed to defer detailed work on mental spaces until Robert returns, focusing instead on refining the semantic constraints and constructions. The team also considered the distinction between focus and topic in information structure, acknowledging the need for further exploration. Finally, they planned to meet again on Thursday at 1 PM to review and refine the thesis proposal and related work. \\

\midrule
\textbf{SumCoT Summary}\\
\midrule

The discussion revolves around the development of a semantic formalism and the integration of mental spaces in linguistics, with researchers like Grad B, Grad F, Grad A, Grad E, Grad D, and Professor C collaborating on the topic. Grad A is set to leave on Thursday, and the group plans to meet again at a time to be determined, possibly at 1 PM or 4 PM. The group is working on refining their notation and semantic constraints, acknowledging the need for further discussion and adjustments. The result is an evolving framework that aims to capture both form and meaning, with a focus on constructions, discourse context, and the potential use of mental spaces for modeling linguistic phenomena. \\

\midrule
\textbf{Highlight-driven planning (End-to-end)} \\
\midrule

The meeting discusses the development of a semantic specification and construction formalism, highlighting the need for a notation that can represent both form and meaning. There is a focus on the distinction between semantic and formal constraints, as well as the use of terms like "designate" and "profile" in the context of mental spaces. The group acknowledges the complexity of handling information structure, including focus and topic, and agrees to defer detailed work on these aspects. They also plan to proceed with defining the scope of their work, focusing on specific domains and using feedback to refine their approach.\\

\midrule
\textbf{Highlight-driven planning (Two-step, ContextCite)}\\
\midrule
The group discussed the revised semantic specification and construction formalism for mental spaces, emphasizing the need to distinguish between form and meaning types. They acknowledged the complexity of mental spaces and decided to defer detailed work until Robert returns, focusing instead on defining the semantic constraints and constructions. The team agreed to prioritize getting the current framework right before considering more abstract or inheritance-based structures. They also noted the importance of handling discourse elements like focus and topic, and mentioned that the notation should allow for flexibility in capturing these aspects.\\

\midrule
\textbf{Highlight-driven planning (Two-step, Generative)}\\
\midrule

The meeting discussed the revised semantic specification and construction formalism, noting that the current notation is similar to previous versions with minimal changes. The group acknowledged the need to clarify terminology, such as "semantic constraints" and "designates," and agreed to defer further decisions on these terms. They also addressed the integration of mental spaces and discourse context, emphasizing the importance of focusing on the core aspects of the formalism before tackling more complex issues. Action items included refining the notation, discussing the semantic side of constructions, and planning a follow-up meeting to review progress and gather feedback. \\

\bottomrule 

\end{xtabular}}

\caption{Comparison between different summaries generated by \texttt{Qwen3-8B} model for a random sample from QMSum validation set. Summaries produced using highlight-driven planning are guided by 30 sentences.  }
\label{tab:qmsum_sample4}%
    
\end{table*}

\begin{table*}[t]

\centering
\footnotesize
  \renewcommand\arraystretch{1}
  \setlength{\tabcolsep}{2.5mm}{

\begin{tabular}{p{1.98\columnwidth}}

\toprule
\textbf{Source Document}\\
\midrule

Industrial Designer: so these are the different options that we have . Okay . So the batteries , I'll start with the battery , right ?

Project Manager: Mm-hmm .

Industrial Designer: So they can be simple which is like uh the normal batteries in uh our {disfmarker} uh the cells , yeah ?

Project Manager: Yeah .

Industrial Designer: Uh thes these are the kind {disfmarker} different kind of batteries that the company makes , right ? So . And dynamos . Um {vocalsound}

Marketing: Does that mean like a wind-up one ?

Industrial Designer: yeah , yeah .

Marketing: {vocalsound} A wind-up remote .

[...]

Industrial Designer: on pressing this button I {disfmarker} a circuit completes , the information goes to the chip , which is somewhere here and the chip that tra then translates the code into an infra infrared radiation , which goes goes out through there . {vocalsound} So uh the important point that I read over the website was uh that the configurations of these printed circuit circuit boards uh are quite cheap to make , you can ge get them printed as you want to ,

[...]

Industrial Designer: Yeah .

Marketing: It it depends on the whole ergonomics of it , you know , it's like how you put your hands so y it's the least movement basically .
Industrial Designer: Yeah . Yeah , singe single side curved or double side curved does not say too much , does it ?

[...]

Industrial Designer: Or or curved at one end and flat on the top , because I I'm not sure if it is flat on both both the sides , then ho how much easy would it be to reach for buttons , etcetera . Um {disfmarker}
Marketing: You have to have a certain element of flatness , I think .

[...]

User Interface: Okay . So um I thought um I would also include the definition of user interface um so it's the aspects of a of of a computer system or programme which can be seen uh by the user um and and which {disfmarker} uh the mechanisms that the user uses to control its operation and input data . So this would p includes things like shape and size and buttons and um voice recognition as well , and colour , and so on .

[...]

User Interface: Um um the method I {vocalsound} employed this time was {vocalsound} a again having a look to related products and mainly on the internet and then {vocalsound} um {vocalsound} analyse them uh from the point of view of user fen friendliness and {vocalsound} also um {vocalsound} whether their appearance was was pleasant . Um {vocalsound} and then um {vocalsound} this uh this um {vocalsound} this can help us to decide which features we want to incorporate in our product .

Project Manager: Mm-hmm .

[...]
Marketing: Imagine it would be inside .

[...]
User Interface: Mm-hmm .

Marketing: Cool , I'm gonna go and sit on my own .

Project Manager: Y ah nobody wants to talk to you .
\\

\midrule
\textbf{Highlights extracted by ContextCite attribution}\\
\midrule

1. Project Manager: Mm-hmm .

2. Yeah , singe single side curved or double side curved does not say too much , does it ?

3. Marketing: Cool , I'm gonna go and sit on my own .

4. Marketing: You have to have a certain element of flatness , I think .

5. Project Manager: Yeah .

6. Project Manager: Mm-hmm .

7. Marketing: Imagine it would be inside .

[...]\\

\midrule
\textbf{Generated highlights}\\
\midrule

1. The chip translates the code into infrared radiation.

2. The Industrial Designer noted that printed circuit boards are cheap to make.

3. The remote can use different types of batteries, including simple, dynamo, and kinetic energy.

4. Marketing suggested a wind-up remote as an idea for kinetic energy.

5. The User Interface Designer mentioned the definition of user interface.

6. The User Interface Designer analyzed products based on user friendliness and appearance.

[...] \\

\bottomrule 

\end{tabular}}

\caption{Comparison between highlight sentences based on ContextCite attribution and highlights generated by LLMs on QMSum validation set}
\label{tab:attr_sample}%

\end{table*}

\begin{table*}[t]

\centering
\footnotesize
  \renewcommand\arraystretch{1}
  \setlength{\tabcolsep}{2.5mm}{

\begin{tabular}{p{1.98\columnwidth}}

\toprule
\textbf{Source Document}\\
\midrule
Background
\\
\\
What is the U.S. International Development Finance Corporation (IDFC)?

The IDFC is authorized by statute to be a "wholly owned Government corporatio n ... under the foreign policy guidance of the Secretary of State in the executive branch. Its purpose is to "mobilize and facilitate the participation of private sector capital and skills in the economic development" of developing and transition countries, in order to complement U.S. development assistance objectives and foreign policy interests (§1412).

[...]

What existing agency functions are consolidated?

[...] The IDFC's authorities would expand beyond OPIC's existing authorities to make loans and guarantees and issue insurance or reinsurance. They would also include the authority to take minority equity positions in investments, subject to limitations. In addition, unlike OPIC, the IDFC would be able to issue loans in local currency. [...] In addition, the IDFC would have the authority to conduct feasibility studies on proposed investment projects (with cost sharing) and provide technical assistance.

What is development finance?

[...] Development finance generally is targeted toward promoting economic development by supporting foreign direct investment (FDI) in underserved types of projects, regions, and countries; undercapitalized sectors; and countries with viable project environments but low credit ratings.

IDFC Organizational Structure and Management

While the IDFC authorized by the BUILD Act has yet to be established, and some implementation questions remain, the act detailed many aspects of how the new entity should be structured, managed, and overseen by Congress. This section discusses the BUILD Act provisions that describe how the new IDFC is expected to function once established.

[...]

How will the IDFC be managed?

The BUILD Act establishes a Board of Directors ("Board"), a Chief Executive Officer (CEO), a Deputy Chief Executive Officer (Deputy CEO), a Chief Risk Officer, a Chief Development Officer, and any other officers as the Board may determine, to manage the IDFC (§1413(a)).

[...]

What will be the responsibilities of the officers?

The BUILD Act establishes four officers for IDFC management.

[...]
\\
\\
IDFC Operations

What financial authorities and tools will the IDFC have?

[...] The BUILD Act also gives the IDFC authority to establish new enterprise funds. It has been argued, however, that the IDFC's authority to conduct equity investment would make enterprise funds unnecessary.
\\

\midrule
\textbf{Highlights extracted by LexRank}\\
\midrule
1. What is the U.S. International Development Finance Corporation (IDFC)?

2. IDFC Organizational Structure and Management

3. How will the IDFC be managed?

4. The BUILD Act establishes four officers for IDFC management.

5. IDFC Operations

6. What financial authorities and tools will the IDFC have?

7. The BUILD Act also gives the IDFC authority to establish new enterprise funds

[...]\\

\midrule
\textbf{Generated highlights}\\
\midrule

1. The U.S. International Development Finance Corporation (IDFC) is a "wholly owned Government corporation" under the foreign policy guidance of the Secretary of State.

2. The IDFC's purpose is to "mobilize and facilitate the participation of private sector capital and skills in the economic development" of developing and transition countries.

3. The IDFC's authorities would expand beyond OPIC's existing authorities to make loans and guarantees and issue insurance or reinsurance.

4. The IDFC would also have the authority to take minority equity positions in investments, subject to limitations.

5. The IDFC would have the authority to conduct feasibility studies on proposed investment projects (with cost-sharing) and provide technical assistance.

6. Development finance generally is targeted toward promoting economic development by supporting foreign direct investment (FDI) in underserved types of projects, regions, and countries.

[...] \\

\bottomrule 

\end{tabular}}

\caption{Comparison between highlight sentences extracted by LexRank and highlights generated by LLMs on GovReport validation set}
\label{tab:lexrank_attr_sample}%

\end{table*}

\end{document}